\documentclass[10pt,twocolumn,letterpaper]{article}

\usepackage[pagenumbers]{cvpr} 

\usepackage[dvipsnames]{xcolor}

\definecolor{cvprblue}{rgb}{0.21,0.49,0.74}
\usepackage[pagebackref,breaklinks,colorlinks,citecolor=cvprblue]{hyperref}

\def\paperID{7} 
\def\confName{CVPR}
\def\confYear{2024}

\title{One-Click Upgrade from 2D to 3D: Sandwiched RGB-D Video Compression for Stereoscopic Teleconferencing}

\author{Yueyu Hu$^{1}$ \quad Onur G. Guleryuz$^{2}$ \quad Philip A. Chou \quad Danhang Tang$^{2}$ \quad Jonathan Taylor$^{2}$,\\
Rus Maxham$^{2}$ \quad Yao Wang$^{1}$\\
$^{1}$Tandon School of Engineering, New York University \quad $^{2}$Google LLC\\
{\tt\small \{yyhu, yaowang\}@nyu.edu} \quad {\tt\small pachou@ieee.org} \\ {\tt\small \{oguleryuz, danhangtang, jontaylor, rrrus\}@google.com} \\ 
}

\begin{document}
\twocolumn[{%
\renewcommand\twocolumn[1][]{#1}%
\maketitle
\begin{center}
    \centering
    \captionsetup{type=figure}
    \includegraphics[width=0.9\textwidth]{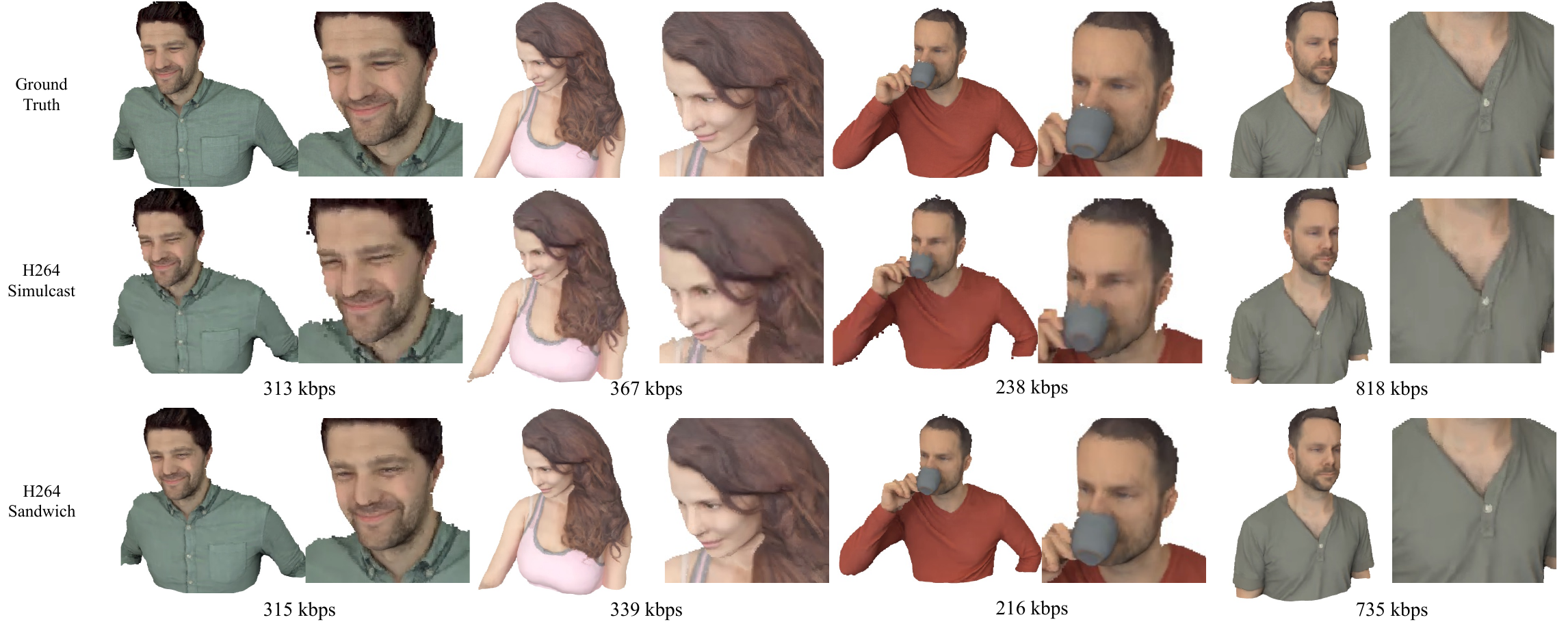}
    \caption{Visualization of decoded 3D representation rendered from a novel view. Results are from both a synthetic and a real-captured dataset. Our method produce a 3D representation maintaining better quality at a lower bit-rate.}
    \label{fig:teaser}
\end{center}%
}]

\begin{abstract}
Stereoscopic video conferencing is still challenging due to the need to compress stereo RGB-D video in real-time. Though hardware implementations of standard video codecs such as H.264 / AVC and HEVC are widely available, they are not designed for stereoscopic videos and suffer from reduced quality and performance. Specific multiview or 3D extensions of these codecs are complex and lack efficient implementations. In this paper, we propose a new approach to upgrade a 2D video codec to support stereo RGB-D video compression, by wrapping it with a neural pre- and post-processor pair. The neural networks are end-to-end trained with an image codec proxy, and shown to work with a more sophisticated video codec. We also propose a geometry-aware loss function to improve rendering quality. We train the neural pre- and post-processors on a synthetic 4D people dataset, and evaluate it on both synthetic and real-captured stereo RGB-D videos. Experimental results show that the neural networks generalize well to unseen data and work out-of-box with various video codecs. Our approach saves about 30\% bit-rate compared to a conventional video coding scheme and MV-HEVC at the same level of rendering quality from a novel view, without the need of a task-specific hardware upgrade.

\end{abstract}    

\section{Introduction}
We aim towards interactive stereoscopic video conferencing, a promising technology that can bring people closer together by enabling them to interact with each other in a natural way. Among various approaches to stereoscopic telepresence~\cite{maimone2012enhanced,zhang2013viewport,jones2009achieving}, we study a typical system that uses stereo videos with color (RGB) images and depth (D) maps to represent a 3D scene. In this system, the server sends the stereo RGB-D videos to the client, allowing it to view the 3D scene from different angles. Here we address the critical compression problem in the remote conferencing scenario that requires low latency and low complexity.

The key idea of compression is to reduce redundancy in the signal. In our scenario, a frame in a stereo RGB-D video is composed of two color-maps and two single-channel depth-maps from two views, respectively. Since the RGB-D videos are two views of one scene captured by two cameras, they have temporal redundancy, cross-view redundancy, and color-depth redundancy that we aim to reduce in our compression scheme. We also consider bit-allocation between the two views and between color and depth.
 

Our compression scheme is inspired by recent works~\cite{guleryuz2021sandwiched,guleryuz2022sandwiched} developing a new architecture called \textit{sandwiched video codec}. In this architecture, a standard video codec is wrapped by a pair of pre- and post-processor built with convolutional neural networks. The neural networks enhance the standard codec, and can further extend its capability to enable high-resolution, high-dynamic-range coding~\cite{guleryuz2022sandwiched}.

In this work, we further explore to make a conventional video codec capable of delivering 3D visual content. Our stereo RGB-D video codec has the following advantages: 1) We lift the complex compression workload onto an optimally engineered video codec, which is usually efficiently implemented in hardware. It keeps our system efficient while offering great compression ratio. 2) The neural processor pair learns to conduct smart bit-allocation and redundancy reduction, greatly alleviating the bandwidth pressure; 3) In contrast to existing standardized multi-view and 3D video codecs~\cite{tech2015overview}, our method does not require further changes made to the hardware implementation. The proposed scheme thus delivers high rate-distortion (R-D) performance and low latency at the same time. As shown in Fig.~\ref{fig:teaser}, we provide better 3D rendering quality at a lower bit-rate compared to a conventional video codec, and this is achieved at a similar latency. This is especially important for the video conferencing scenario, since it requires real-time encoding and decoding.


To achieve the compression capability, we develop multiple techniques to further save bit-rates from a conventional simulcast video coding setting, and to maintain good rendering quality.
Given the nature of stereo RGB-D signal, one of our key innovation is to allow the encoder to process the color images and depth maps jointly. This allows the encoder to analyze the content and perform bit-allocation for different regions with different geometric and content texture characteristics. This design also facilitates cross-view redundancy reduction, leading to more compact representation.
In addition, we propose a disparity warping-based distortion loss function to improve the depth quality in the context of rendering. This loss function enables the pre- and post-processor to allocate more bits to critical areas where the rendering is sensitive to depth errors.

We evaluate our method with H264 / AVC~\cite{wiegand2003overview} and HEVC~\cite{sullivan2012overview} on an unseen synthetic testing dataset and a real captured dataset built from \cite{guo2019relightables}. Experimental results show that our method reduces the bit-rate over existing solutions by 29.3\%, while maintaining the same level of rendering quality. 
We also demonstrate that jointly processing color and depth (detailed in Fig.~\ref{fig:preprocessor}), and employing the warping-based depth loss (detailed in Fig.~\ref{fig:loss}), both benefit the rate-rendering-distortion performance.


Our contributions are summarized as follows:
\begin{itemize}
    \item We propose a novel sandwiched video compression scheme for stereo RGB-D video streaming, which is shown to reduce the bit-rate over existing solutions by 29.3\%, while maintaining the same level of rendering quality. The proposed scheme can be practically deployed over existing video streaming infrastructure.
    \item We develop a disparity warping-based distortion loss function to improve the depth quality in the context of rendering, which enables the pre- and post-processor to allocate more bits to critical areas where the rendering is sensitive to depth errors.
    \item We propose to transform depth maps into world-space coordinates to facilitate stereo alignment, and show that the preprocessor can benefit from it to improve the R-D performance especially at the higher bit-rates.
\end{itemize}

\begin{figure*}[t]
  \centering
  \includegraphics[width=0.95\linewidth]{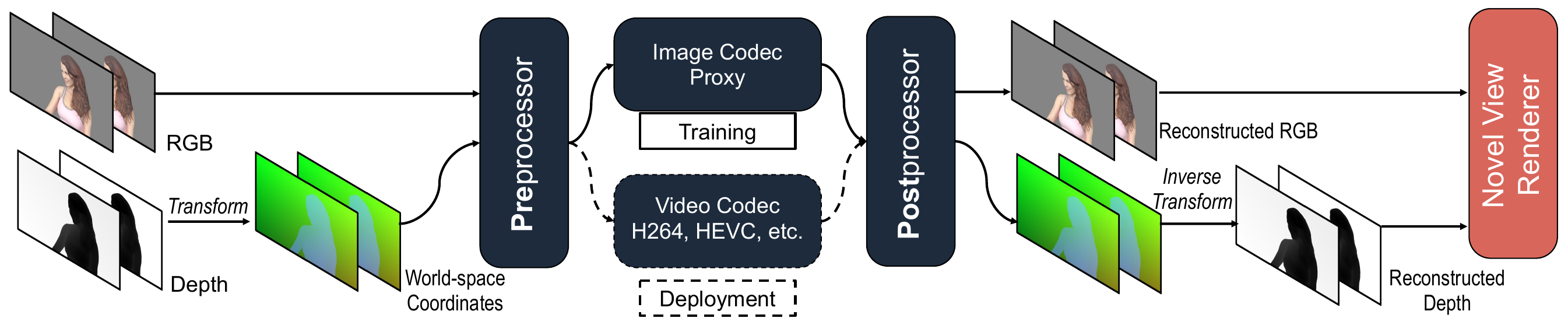}
  \caption{Sandwiched compression scheme for stereo RGB-D video streaming.}
  \label{fig:framework}
  \vspace{-4mm}
\end{figure*}

\section{Related Works}

\subsection{Multi-View Video Coding}
Coding images and videos of one scene captured by multiple cameras has attracted broad interest in the past decades~\cite{perkins1992data,lukacs1986predictive,varadarajan2012rgb,liu2015hybrid}. 
The core idea of recent achievements in multi-view video coding based on standard codecs~\cite{vetro2011overview,tech2015overview} is to utilize the established motion-compensation tools for cross-view predictive coding. 
This technique usually does not require significant change in the decoder structure.  However, since the encoding process is complicated, there is a lack of an efficient implementation of a multi-view video encoder, and there are no practical solutions for low-delay encoding. Therefore, such techniques are still far from practical use in stereo video conferencing. Specifically, 3D-HEVC encodes both the depth map and color frames and shows better performance, but it requires major changes in the encoder and decoder pipeline to allow geometry-specific compensation, making it more challenging to deploy on consumer devices.

In our work, we propose a new scheme for cross-view and cross-modality coding that is more practical for real-world use than existing methods. Our scheme offloads the burden of cross-view and cross-modality coding to the pre- and post-processor, and adopts a general purpose video codec to handle temporal redundancy. This makes our scheme more efficient and easier to implement in hardware, while achieving a great improvement in rate-distortion performance.

\subsection{Learned Stereo Image and Video Compression}
With the proliferation of learned image and video codecs~\cite{balle2020nonlinear,hu2021learning,liu2020deep,li2022hybrid,lu2019dvc,hu2020improving}, there has been a growing interest in using learning-based methods for stereo / volumetric image and video compression. In \cite{lei2022deep,wodlinger2022sasic,liu2019dsic,chen2022exploiting,volokitin2022neural}, neural network-based stereo image compression schemes are proposed. In \cite{lei2022deep,liu2019dsic} the network architecture allows information sharing and compensation between the feature maps from two views, while in \cite{wodlinger2022sasic} bit-rate is reduced by coding a second view conditioned on the feature maps from a previously coded view. The idea of feature representation compensation has also been extended to stereo video coding~\cite{chen2022lsvc}, where the neural network conducts disparity and motion compensation simultaneously.

In direct contrast to these series of work, our work considers \textbf{color and geometry} joint coding and targets rendering quality. We show that it is beneficial to have the scheme consider color and geometry together to optimize rendering quality. Nevertheless, our scheme is further capable of achieving higher performance or higher efficiency given progress made to general video coding. For example, our pre- and post-processors can be combined with a state-of-the-art neural video codec in a plug-and-play fashion and achieve higher rate-distortion performance. Therefore, our scheme has more flexibility than existing learned stereo video compression works.

\section{Method}

\subsection{Sandwiched Video Compression Scheme}




We aim to design a practical scheme to efficiently deliver stereo RGB-D video. The system should have the following desired characteristics:
1) \textbf{Simulcast support}. This means that different views (\textit{i.e.}, left and right) and different modalities (\textit{i.e.}, RGB and depth) are transmitted simultaneously. This allows the system to reduce latency in a real-time video conferencing scenario.
2) \textbf{Smart bit-allocation}. The system should should efficiently allocate bits between views and modalities to achieve an optimal rate-distortion tradeoff.
3) \textbf{Cross-view/cross-modality redundancy reduction}. The system should be able to reduce cross-view/cross-modality redundancy for more efficient coding.
4) \textbf{Generalization to different camera configurations}. The system should be able to generalize to different camera configurations, such as different baselines and different resolutions.

To achieve these goals, we build the stereo RGB-D video compression scheme by \textit{sandwiching} a traditional video codec with preprocessing and post-processing neural networks, shown in Fig.~\ref{fig:framework}. 
In contrast to a video simulcast baseline that directly codes the RGB and depth frames independently, we employ a neural network preprocessor that takes the color and geometry as input and generates neural codes, in the format of video frames, to be coded by the video codecs. The transmitted neural codes are postprocessed by the neural postprocessor to reconstruct the color and geometry for client-side rendering. The pre- and post-processor are jointly trained neural networks. To make them generate neural codes suitable for conventional video coding, we employ an image codec proxy to enforce a rate-distortion constraint onto the neural codes. With the proxy, the preprocessor is trained to generate neural codes as color images that can be efficiently compressed by a real image codec. We show that such neural code generation generalize to other transform coding based hybrid video codecs (\textit{i.e.} H264 / AVC and HEVC).

Our pre- and post-processor handle each frame in the video independently. Each stereo RGB-D frame includes RGB and depth maps in the left and right views. Since the left and right depth maps are defined in the left and right camera coordinate system, respectively, to make the preprocessor better explore cross-view redundancy, we apply a camera-to-world coordinate transformation on each of the depth maps, making the geometric inputs to the preprocessor lie in the canonical coordinate system (\textit{i.e.} each pixel is endowed three coordinate values defined on a common 3D coordinate system). The preprocessor takes the 12-channel images as input, and produces $K$-channel \textit{bottleneck} neural code images. After lossy encoding and decoding, either by the image proxy at the training time or a video codec in actual deployment, the post-processor converts the compressed bottleneck neural code images to the reconstructed RGB images and canonical space coordinates. The camera-specific world-to-camera transformations are applied to the coordinates to produce the depth maps, which together with the color images serve the novel view rendering.

We mainly choose $K = 12$ in our experimental settings to provide enough flexibility for neural network optimization. Since 12 channels can be organized into 4 video streams, we are not coding extra motion vectors compared to a simulcast codec. Through end-to-end learning, the neural networks learn to occupy adequate number of channels and zero out certain channels to save bit-rates in a tight bandwidth budget. Nevertheless, one can set a smaller $K$ for a specific scenario that requires a lower number of video streams.

\subsection{Preprocessor}

\begin{figure}[t]
\centering
  \includegraphics[width=0.9\linewidth]{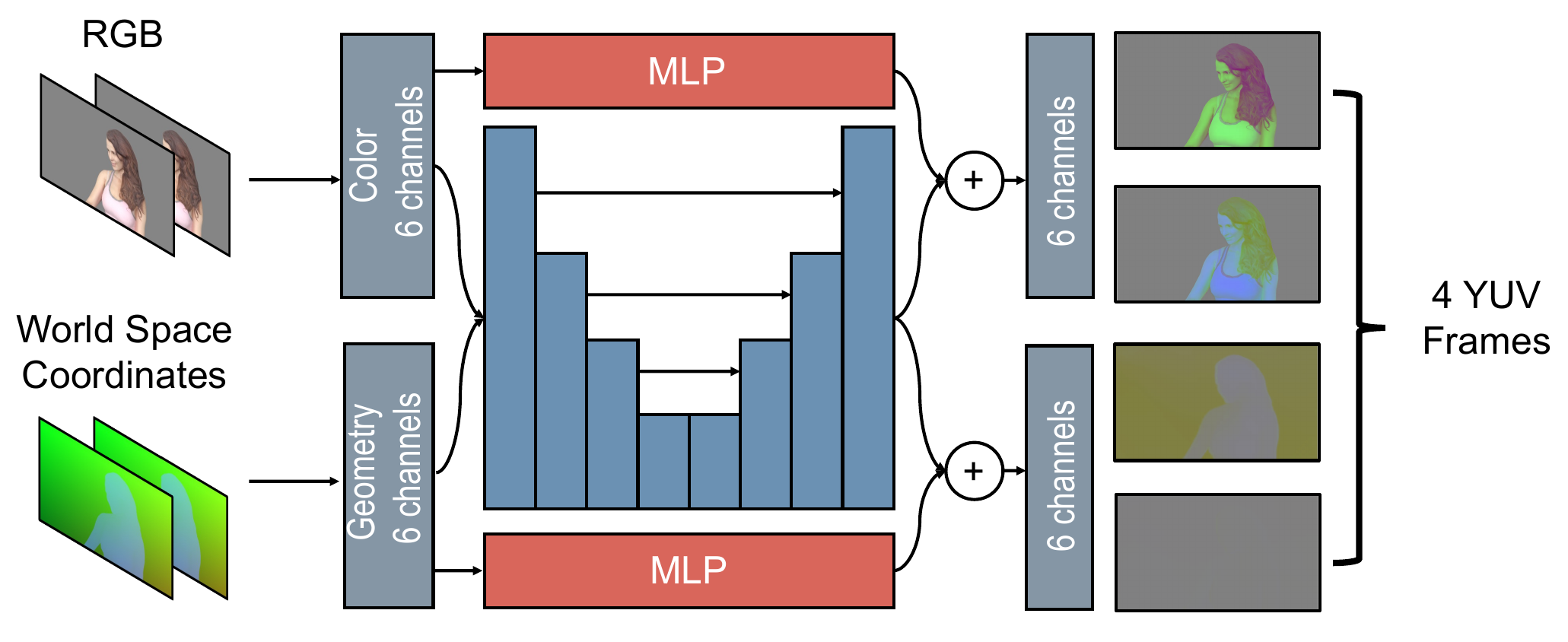}
  \caption{Neural network architecture of the preprocessor.}
  \label{fig:preprocessor}
  \vspace{-3mm}
\end{figure}

The neural network-based pre- and post-processors are the key components in the proposed scheme to conduct smart bit-allocation and redundancy removal across views and modalities. We adopt the architecture design from other sandwiched codec schemes~\cite{guleryuz2021sandwiched}, which is composed of a U-Net and a multilayer perceptron (MLP) or equivalently multiple 1x1 2D convolutional layers. We specifically design the preprocessor architecture for multi-modality processing. The structure is illustrated in Fig.~\ref{fig:preprocessor}.

It is discussed in \cite{isik2023sandwiched} that the MLP usually does the per-pixel color space conversion, while the U-Nets add or remove modulation patterns. In our scenario, the color space transformation and the geometry transformation have different physical meanings. Therefore it is better to handle them separately. In our proposed design, two separate MLPs process the color and geometry maps, each in two views and 6 channels, respectively. Each of the MLP generates 6-channel neural code maps from the 6-channel input. To allow further cross-modality bit-allocation and rate reduction, a U-Net takes the concatenated RGB and coordinate maps as input, and generate 12-channel feature maps. These feature maps are split by half and added to the two groups of feature maps generated by the two MLP, respectively. The preprocessor produces feature maps (neural codes) in two groups, each with 6 channels. When using a video codec to encode the neural codes in each group, we organize them into 2 YUV 4:4:4 sequences. We choose the 2 channels that are highest in pixel variance as the luma component, and the other 4 channels are organized as the chroma channels. This is to best utilize motion estimation and compensation tools in modern video codecs.

\subsection{Loss Function}


We train our models towards the optimal rate-distortion tradeoff as follows,
\begin{equation}
    \begin{split}
        L_{RD} = D + \gamma R,
    \end{split}
    \label{eq:rdloss}
\end{equation}
where the bit-rate term $R$ is estimated by a JPEG proxy over all neural code maps, following the definition in \cite{guleryuz2021sandwiched}. We train different models with different values of $\gamma$ to cover different ranges of bit-rate.

\begin{figure}[t]
\centering
  \includegraphics[width=0.9\linewidth]{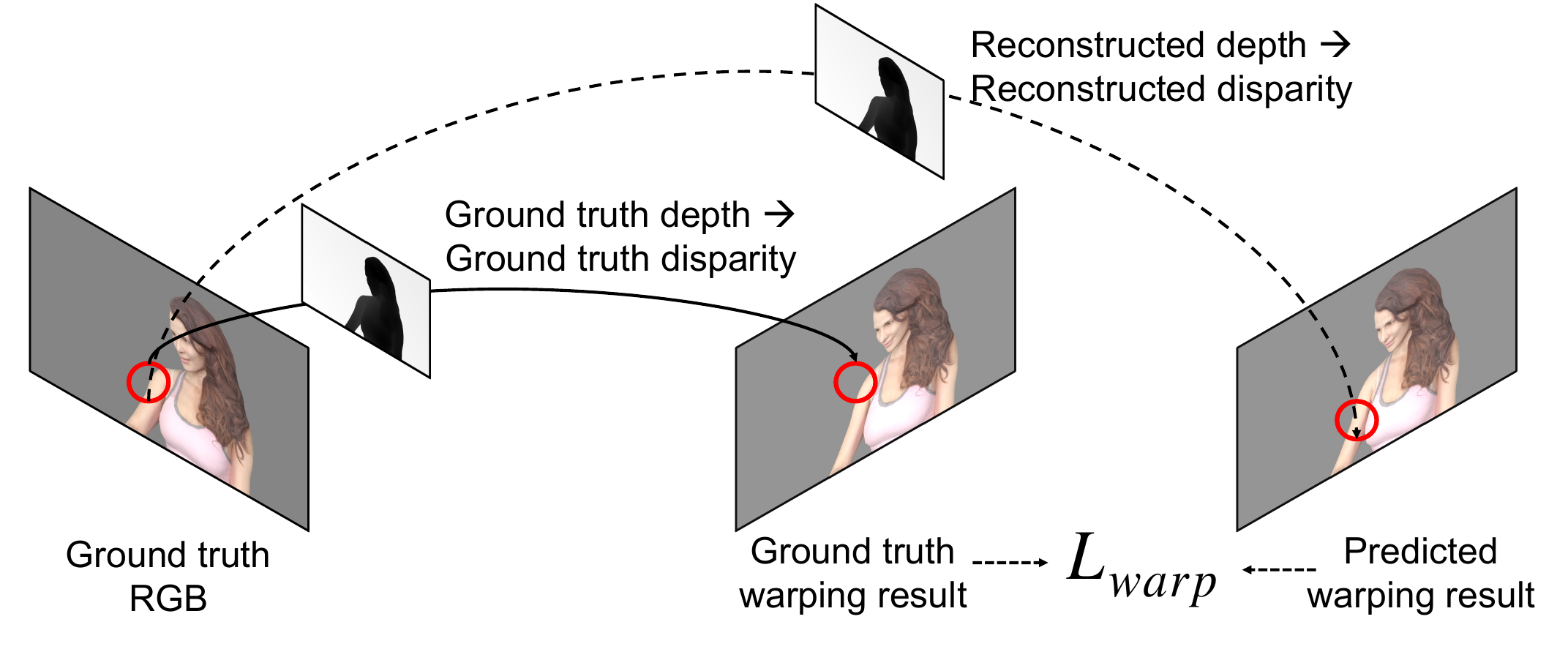}
  \caption{Warping-based depth distortion loss function.}
  \label{fig:loss}
    \vspace{-3mm}
\end{figure}

We aim to find a distortion loss that well approximates the actual novel-view rendering distortion to enables effective rate-allocation between color and geometry, and at the same time is efficient for training. The distortion term in Eq.~(\ref{eq:rdloss}) is formulated as weighted sum of the color image distortion and depth distortion, where we specifically design the depth distortion, illustrated in Fig.~\ref{fig:loss}. We are given a ground truth depth map $Z$ and the reconstructed depth map $\hat{Z}$, both corresponding to the ground truth color image $I$. In the following, without loss of generality, we take the left view as an example. Terminology can be changed to the other view respectively.

Since we also have the camera configuration, we can assume the image $I$ is the left image and calculate the disparity $d$ between the left and right camera from $Z$ and $\hat{Z}$ as,
\begin{equation}
    \begin{split}
        d = f(Z; \phi), \text{ and  }
        \hat{d} = f(\hat{Z}; \phi),
    \end{split}
\end{equation}
where $\phi$ is the set of intrinsic and extrinsic parameters of the stereo cameras. With the disparity, we are able to warp the color image pixels in the left image $I$ to the right image $J$. The warping in effects acts like rendering a novel view of the scene represented by the color images and depth maps. We calculate the difference between the warped images as the warping loss, as,
\begin{equation}
    \begin{split}
        & L_{\text{warp}} = ||J - \hat{J}||_2, \\
        & J = f_{\text{warp}}(I, d), \text{ and } \hat{J} = f_{\text{warp}}(I, \hat{d}).
    \end{split}
\end{equation}
The warping function $y = f_{\text{warp}}(x, u)$ generates a new image $y$ by resampling from the input image $x$, following a dense flow $u$ and apply bilinear interpolation for fractional coordinates.

The distortion term in Eq.~(\ref{eq:rdloss}) is then formulated as weighted sum of the color image distortion, warping-based depth distortion, and depth map MSE, namely,
\begin{equation}
    \begin{split}
        D = f_d(I, \hat{I}) + \alpha L_{\text{warp}} + \beta ||Z - \hat{Z}||_2.
    \end{split}
    \label{eq:loss}
\end{equation}
Typical differentiable distortion functions $f_d$ for color images include MSE, SSIM, and LPIPS~\cite{zhang2018unreasonable}. $\alpha$ and $\beta$ are hyperparameters to control the strength of depth fidelity regularization.
\section{Experiments}

\subsection{Settings}

Our method is designed for any 3D streaming system that transmits stereo RGB-D videos. In the following, we describe one of the design options.

The prospective teleconferencing system consists of an up-link and a down-link. In the up-link, the sender captures multi-view images of a person from multiple calibrated cameras, and send them to a server as video streams. With the calibrated multi-view images, the server conducts a 3D reconstruction process and build a high quality mesh. The server then renders stereo interpupillary distance (IPD) RGB-D images out of the mesh. With the down-link, the server streams the IPD stereo RGB-D video to the receiver. The color images will allow stereo perception at the receiver. The depth images further support more efficient interactive novel-view rendering and therefore provide more degrees of viewing freedom.

In this work, we particularly focus on the down-link compression problem. To make the system practical, the latency of the down-link compression system should be small. With this in mind, we specifically design our system to be standard hardware compatible. The decoder of our system can be practically implemented on a modern graphics card with hardware video decoder and GPU. The bit-stream produced by our system can be first decoded by an on-board standard decoder, and then directly processed by the GPU. Therefore, our system can be potentially practically deployed on existing commercial hardware and offers the upgrade from 2D to 3D video conferencing experience.

We simulate this VR video conferencing scenario to evaluate our compression scheme. In this scenario, we focus on the upper body of a human subject performing different actions. We assume that the background is static and in simple color. We evaluate the multi-novel-view rendering quality at the client side with different rate constraints.

\subsubsection{Dataset}

We build a synthetic dataset of 4D people to train and evaluate the neural network-based pre- and post-processor. We render 3D meshes of people performing actions from two cameras placed in an interpupillary distance (IPD) setting. The rendering results include the color image and depth map from the perspective of each camera. The dataset consists of RGB-D stereo videos from 44 subjects, each performing 5 randomly chosen actions. Each video in the dataset shows a figure performing in front of a gray backdrop. Among the 44 subjects, 9 are randomly selected for held-out testing and the remaining 35 subjects are for training. All 3D models are randomly rotated (by -30, 0 or 30 degrees). Note that in the 10 testing sequences, \#2 and \#8 are from the same subject but the action and the distance of the figure to the camera differs.

To demonstrate that our method, although trained on a synthetic dataset, can generalize to real world scenario, we also evaluate our method on \textit{The Relightables}~\cite{guo2019relightables} dataset. This dataset contains relightable meshes from 7 real-captured human subjects in actions. We generate 10 testing stereo RGB-D video sequences with the same IPD camera setting from captured subjects in this dataset. We only use these data to test our models, which are trained only with synthetic data.

\subsubsection{Baselines}
We compare our method to existing solutions based on video codecs, including H264 / AVC, HEVC in the simulcast setting and MV-HEVC that is specifically designed for multi-view videos. Since we propose to wrap a video codec by a neural network-based pre- and post-processor pair, in experiments we basically compare the \textit{sandwiched} scheme to the video codec itself. Namely, we compare a H264 Sandwich to H264 Simulcast, and HEVC Sandwich to HEVC Simulcast.

\textbf{H264 Simulcast}. The stereo RGBD video is organized as two RGB color videos and two grayscale depth videos. Each of the four is coded independently by H264. We use libx264 with FFmpeg, which takes raw RGB video as input, convert it to YUV 4:4:4 for H264 coding, and decodes and convert to RGB video within FFmpeg. Depth maps are formatted as grayscale videos for direct FFmpeg / libx264 coding.

\textbf{HEVC Simulcast}. We use the reference software HEVC Test Model (HM) 16.3. Since we are targeting video conferencing scenario, we use the low-delay-P YUV 4:4:4 configuration. Same as H264 we convert RGB videos to the YUV 4:4:4 format. For depth, the single-channel depth signal is placed in the Y channel and UV channels are filled with zeros.

\textbf{MV-HEVC}. We use the MV-HEVC reference software in HM version 16.3.  Since MV-HEVC only supports YUV 4:2:0, in addition to the RGB-YUV color conversion, we further down-sample the UV channels. They are upsampled before converted to RGB at the decoder. We code the two color streams in one bit-stream as two views with MV-HEVC. Two Y channels carrying depth maps and UV filled with zeros are coded into the other bit-stream as two views.

\subsubsection{Our Scheme}
We train 6 pre- and post-processors with $\gamma \in \{0.5, 2, 4, 8, 16, 32\}$. We use a JPEG proxy during training. Once trained, the pre- and post-processor work together with H264 and HEVC.

\textbf{H264 Sandwich}. We use the same setting as the baseline H264, except that we treat the neural code directly as YUV 4:4:4 frames to avoid unnecessary color conversion. The neural code given by the preprocessor are organized into 4 three-channel streams as shown in Fig.~\ref{fig:preprocessor}.
To better utilize motion estimation in the sandwiched codec, channels with the highest variance are chosen as the Y channels. The ordering is signaled to the decoder for each video.

\textbf{HEVC Sandwich}. We generate the YUV sequences of neural codes in the same way as H264 Sandwich.
Those sequences are coded with HM, using low-delay-P configuration, and decoded to obtain the reconstructed neural code.

\subsubsection{Metrics}
All methods take stereo RGB-D videos as input, code them into bit-stream, and reconstruct the stereo RGB-D videos. We compare rate-distortion performance among our methods and baselines. We first build a mesh from the decoded depth and color images. Each RGB-D pixel defines the coordinate 
and the color of a vertex, respectively. We then render the mesh using TensorFlow graphics triangle rasterizer~\cite{tfgraphics} in 4 different novel views, with the camera moving in the (-3cm, 3cm) range horizontally. We evaluate the distortion between the novel view rendered from the ground truth RGB-D and the reconstructed one. Since we are using a dummy background, we calculate the PSNR \textit{w.r.t.} valid pixels, \textit{i.e.} we exclude pixels in the gray backdrop. Each R-D point is evaluated by choosing a model trained to a specific $\gamma$ and code with a specific video codec quantization parameter (QP). Since we have 6 different models trained with different values of $\gamma$, we plot R-D curves by first calculating a convex hull from the all R-D points, and average points from different sequences at the same slope. We also report BD-Rate~\cite{bjontegaard2001calculation} which shows the average bit-rate saving at the same level of distortion.

\subsection{Rate-Distortion Performance}

\subsubsection{Synthetic Testing Dataset}

\begin{figure}[t]
    \centering
  \includegraphics[width=0.8\linewidth]{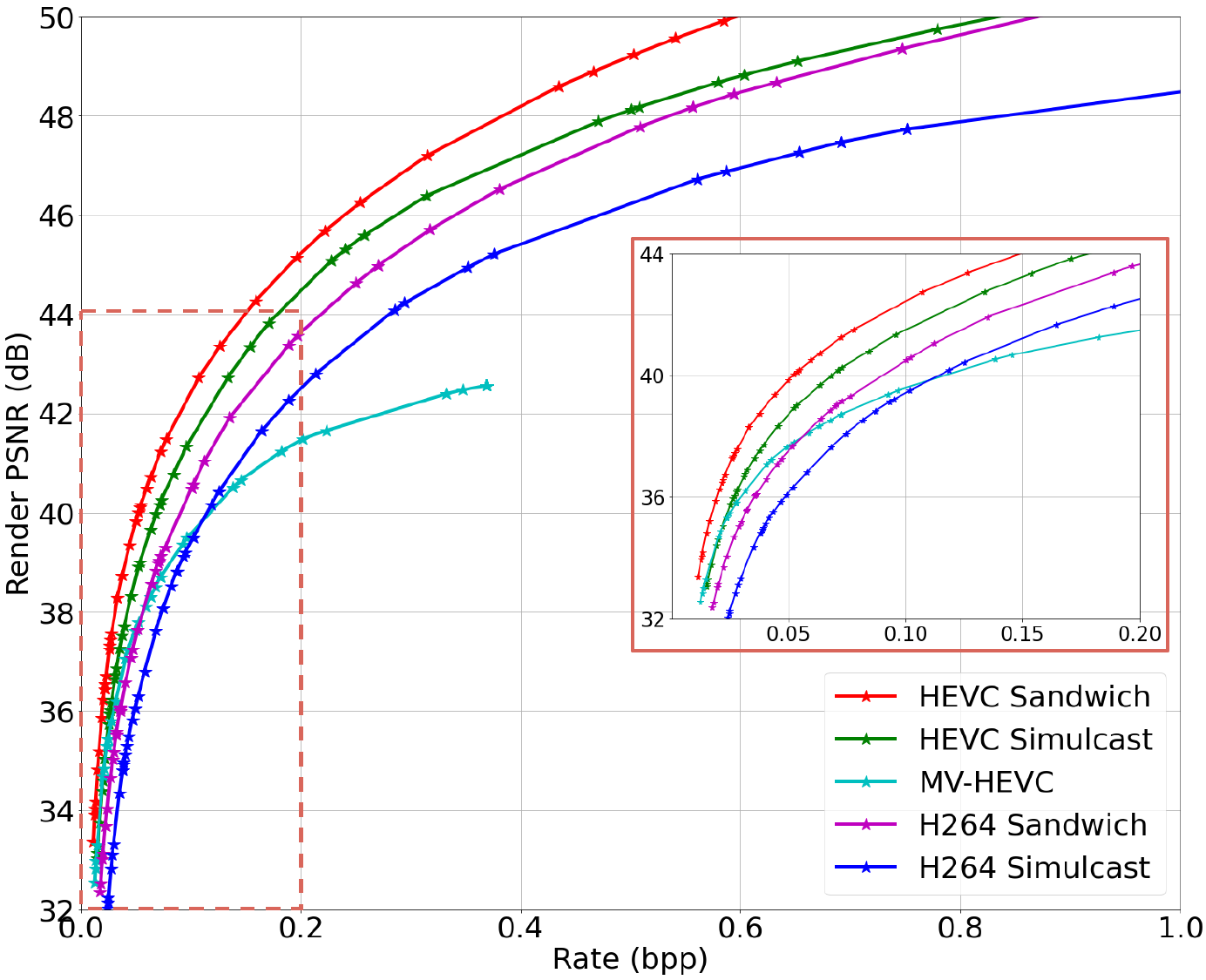}
  \vspace{-3mm}
  \caption{Aggregated rendering rate-distortion curves on the synthetic testing dataset.}
  \label{fig:main_rd}
  \vspace{-3mm}
\end{figure}

\begin{table}[t]%
\caption{Per-sequence BD-Rate on the synthetic testing dataset. Negative numbers indicate bit-rate saving.}
\label{tab:overall_bdrate}
\begin{minipage}{0.45\columnwidth}
\begin{center}
    \footnotesize
\begin{tabular}{c|c}
  \toprule
  Sequence ID & BD-Rate \\
  \midrule
    0 & -36.4 \% \\
    1 & -39.9 \% \\
    2 & -23.0 \% \\
    3 & -37.4 \% \\
    4 & -23.9 \% \\
    5 & -33.7 \% \\
    6 & -33.3 \% \\
    7 & -7.6 \% \\
    8 & -31.1 \% \\
    9 & -26.1 \% \\
    \midrule
    Avg & -29.3 \% \\
  \bottomrule
\end{tabular}
\end{center}
\bigskip\centering

\vspace{-4mm}
Sandwiched / H264
\vspace{-4mm}
\end{minipage}
\begin{minipage}{0.45\columnwidth}
\begin{center}
    \footnotesize
\begin{tabular}{c|c}
  \toprule
  Sequence ID & BD-Rate \\
  \midrule
    0 & -32.5 \% \\
    1 & -40.0 \% \\
    2 & -23.5 \% \\
    3 & -35.2 \% \\
    4 & -20.7 \% \\
    5 & -28.3 \% \\
    6 & -28.0 \% \\
    7 & -10.2 \% \\
    8 & -28.7 \% \\
    9 & -24.3 \% \\
    \midrule
    Avg & -27.1 \% \\
  \bottomrule
\end{tabular}
\end{center}
\bigskip\centering
\vspace{-4mm}

 Sandwiched / HEVC
 \vspace{-4mm}
\end{minipage}
\end{table}%

We first demonstrate the improvement in rendering rate-distortion performance on the synthetic testing dataset. It is shown in Fig.~\ref{fig:main_rd} that our scheme, although trained with simple JPEG proxy, works well with more sophisticated video codecs and brings significant improvements in rate-distortion performance. We also include results from MV-HEVC in the comparison. However, MV-HEVC only supports YUV 4:2:0, making it less competitive at the higher bit-rate region.

Table~\ref{tab:overall_bdrate} shows the quantitative BD-Rate results, calculated \textit{w.r.t.} each testing sequence in the PSNR range 32 to 50 dB. Each table presents the bit-rate reduction given by sandwiching the specified codec (\textit{i.e.} H264 and HEVC) with our pre- and post-processor. As shown, our proposed method steadily achieves over 29.3\% bit-rate saving on most of the sequences with H264. Similar gain is observed with HEVC.

\subsubsection{Real Captured Data}

\begin{figure}[t]
    \centering
  \includegraphics[width=0.9\linewidth]{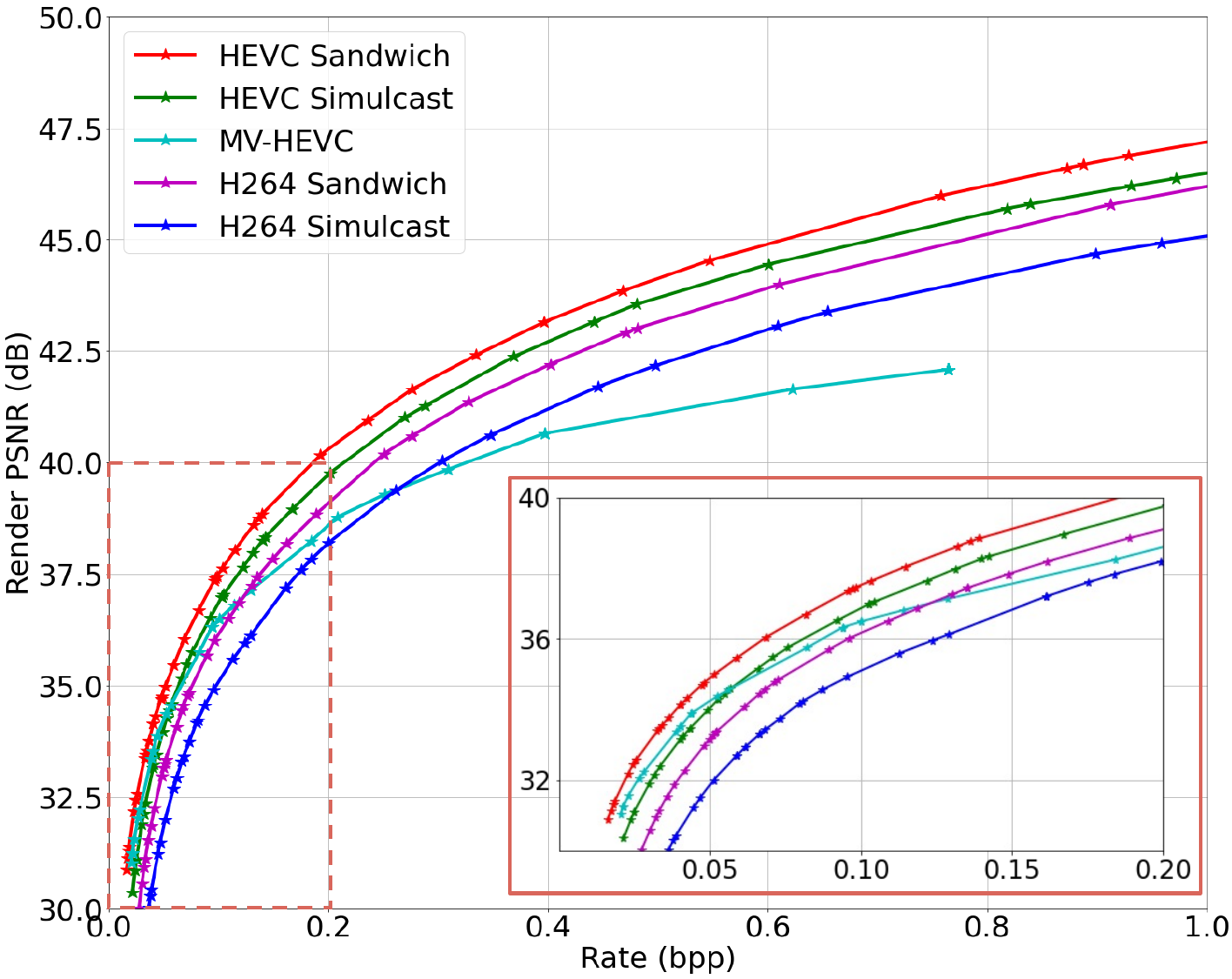}
  \vspace{-3mm}
  \caption{Aggregated rendering rate-distortion curves on \textit{The Relightables} dataset.}
  \label{fig:real_rd}
  \vspace{-4mm}
\end{figure}

To demonstrate that our models generalize well to real use cases, we evaluate our method on sequences obtained from \textit{The Relightables} dataset. As shown by the R-D curves in Fig.~\ref{fig:real_rd}, although the models are trained on synthetic dataset, they generalize well on real captured scenes.

\begin{figure*}[t]
    \centering
    
    \begin{subfigure}[h]{0.24\linewidth}
      \includegraphics[width=1\linewidth]{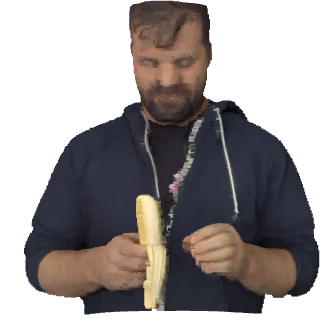}
      \centering
      \caption{H264 648 Kbps}
    \end{subfigure}
    \begin{subfigure}[h]{0.24\linewidth}
      \includegraphics[width=1\linewidth]{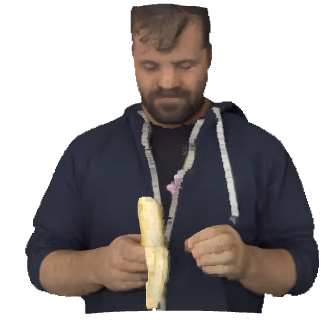}
      \centering
      \caption{Ours 562 Kbps}
    \end{subfigure}
    \begin{subfigure}[h]{0.24\linewidth}
      \includegraphics[width=1\linewidth]{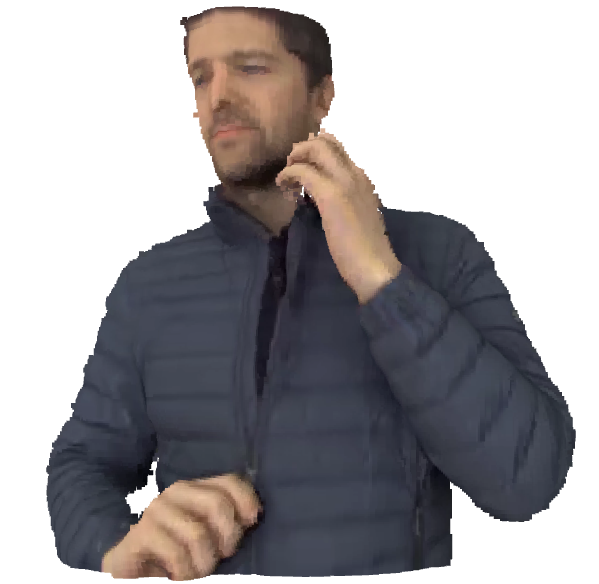}
      \centering
      \caption{H264 820 Kbps}
    \end{subfigure}
    \begin{subfigure}[h]{0.24\linewidth}
      \includegraphics[width=1\linewidth]{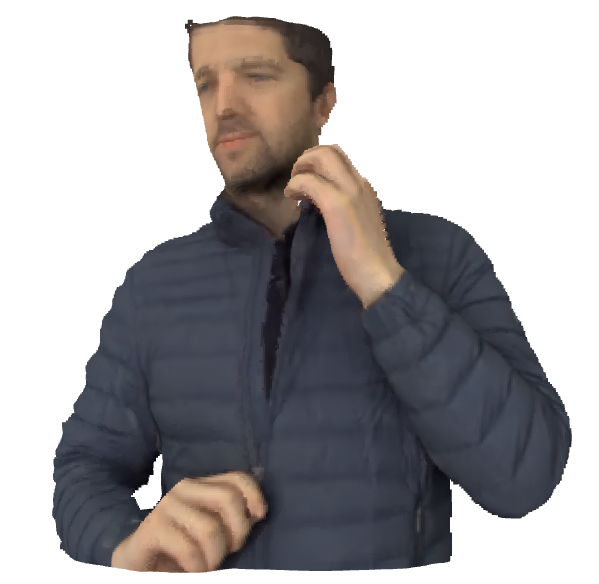}
      \centering
      \caption{Ours 775 Kbps}
    \end{subfigure}
      \caption{Rendering results with signal compressed at moderate bit-rates.}
      \label{fig:render_people_pcloud}
    \end{figure*}

\subsection{Ablation Study}

\subsubsection{Loss Function}

\begin{figure}[t]
    \centering
  \includegraphics[width=0.9\linewidth]{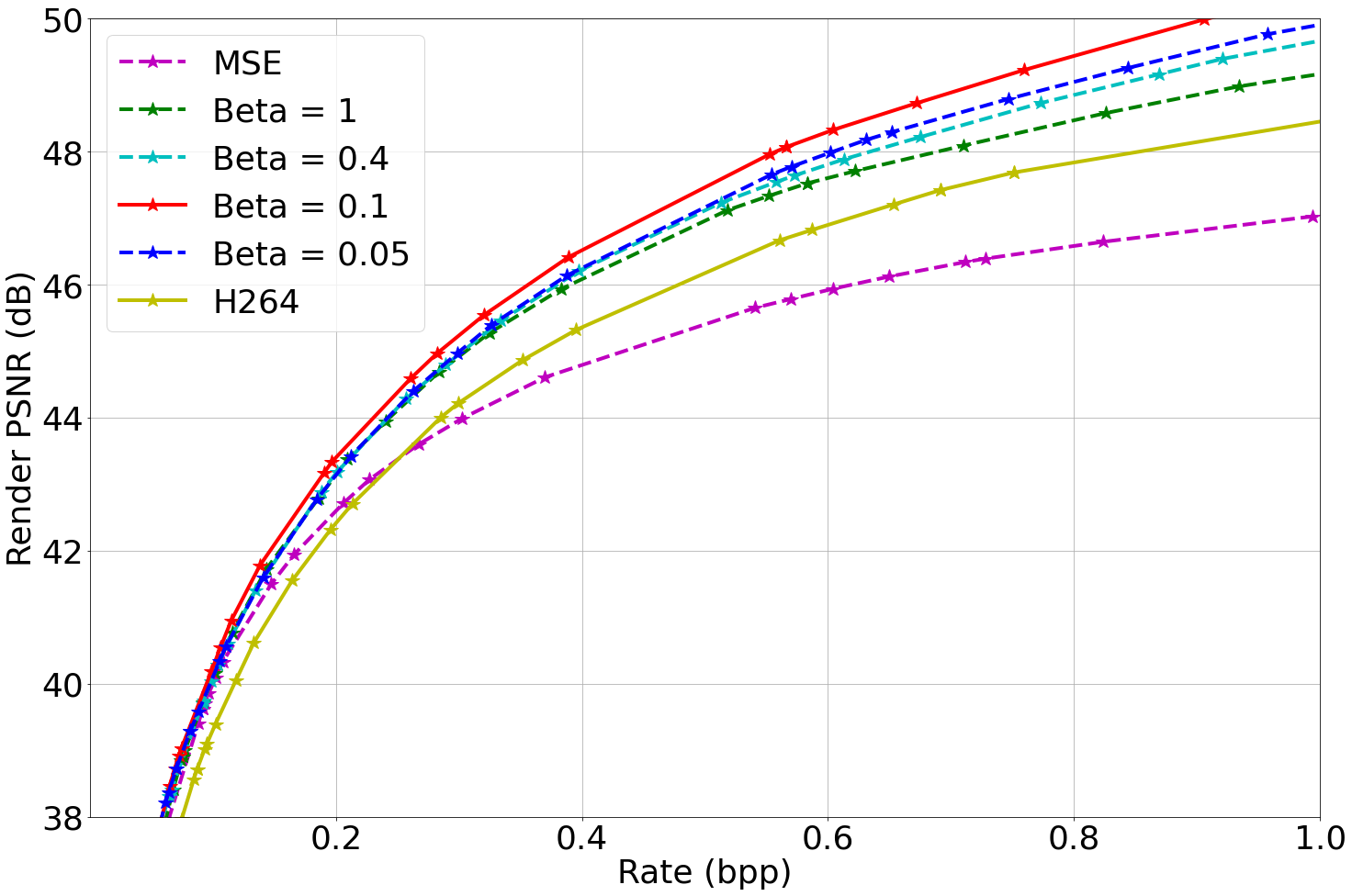}
  
  \vspace{-3mm}
  \caption{Rendering distortion-rate curves of models trained with different loss function hyper parameters.}
  \label{fig:ablation_loss}
\end{figure}

We analyze the impact of distortion term in the loss function by setting different values for $\alpha$ and $\beta$ in Eq.~\ref{eq:loss}. Fig.~\ref{fig:ablation_loss} shows the comparison between different loss function settings, where \textit{MSE} means $\alpha = 0$ and $\beta = 1$. In all other settings when we vary $\beta$, we set $\alpha=1$. We can see that models trained with RGB-D MSE loss under-perform the H264 baseline. By adding the $L_{\text{warp}}$ term in the distortion loss, the rendering performance is overall improved. The results also show that it may be beneficial to reasonably reduce the weight for the depth MSE term when $L_{\text{warp}}$ is used, but the rate-distortion performance is not very sensitive to $\beta$. We use $\alpha = 1$ and $\beta = 0.1$ as the main setting.

\subsubsection{Architecture}

In the proposed scheme, we transform the depth values to the canonical coordinate space, and employ one U-Net with two MLPs to process the color and geometry, respectively.
In this experiment, we study configurations with one U-Net but different numbers of MLPs for the preprocessor, \textit{i.e.}, 
\begin{itemize}
    \item \textbf{Two MLPs}. The main architecture shown in Fig.~\ref{fig:framework}.
    \item \textbf{Four MLPs}. We employ four MLPs in the preprocessor besides the U-Net, for different views (left and right) and different modalities (color and geometries). Each of the MLPs takes 3 channels as input and produce 3 output channels.
    \item \textbf{Single MLP}. One U-Net with one MLP, \textit{i.e.} the default architecture in \cite{guleryuz2021sandwiched}.
    \item \textbf{Geometry: Depth}. The preprocessor takes depth maps as input and does not transform them to the canonical space.
    \item \textbf{Geometry: Camera Coordinates}. Camera-space 3D coordinates constructed from the depth maps are used as the input. The coordinates from the two views are in different coordinate space (\textit{i.e.} the left and right camera space, respectively).
    \item \textbf{Color Only}. We only use the proposed scheme to encode and decode the left and right color maps. The depth maps are coded by H264.
\end{itemize}

To better analyze different settings in different bit-rate ranges, we divide the valid PSNR range into two sub-ranges, \textit{i.e.} lower bit-rate (32 - 42 dB) and higher bit-rate (42 - 52 dB). We evaluate BD-Rate in different ranges for comparison. The results are shown in Table.~\ref{tab:ablation_arch}. We can see that having separate MLPs for the color and geometry greatly benefits the performance, especially for the higher bit-rates. Further splitting the MLP for different views brings a little more gains but also increase the complexity. Hence we basically adopt the two-MLP architecture. It is also shown that transforming coordinates to the canonical space greatly helps improve performance at the higher bit-rates. Finally, by comparing the \textit{color only} scheme to the proposed one that jointly handles color and geometry, we show that in the scenario of stereo RGB-D compression for rendering, it is important to jointly process and compress the color and geometry. Existing methods that handle color and geometry separately may be suboptimal.

\begin{table}%
\caption{Average BD-Rate on the synthetic testing dataset.}
\label{tab:ablation_arch}
\begin{minipage}{1\columnwidth}
\begin{center}
    \footnotesize
\begin{tabular}{c|c|c}
  \toprule
  Bit-Rate Range & Lower & Higher \\
  \midrule
  Two MLPs & -25.0 \% & \textbf{-29.8} \% \\
  Four MLPs & \textbf{-28.8 \%} & -29.0 \% \\
  Single MLP & -22.1 \% & -12.7 \% \\
  Geometry: Depth & -27.4 \% & -20.6 \% \\
  Geometry: Camera Coordinates & -24.1 \% & -20.1 \% \\
  Color Only: & -9.5 \% & -19.3 \% \\
  \bottomrule
\end{tabular}
\end{center}
\bigskip\centering
\end{minipage}
\vspace{-10mm}
\end{table}%

\begin{figure}[t]
    \centering
    \begin{subfigure}[h]{0.48\linewidth}
      \includegraphics[width=1\linewidth]{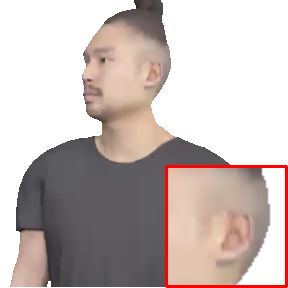}
      \centering
      \caption{H264 110 kbps}
    \end{subfigure}
    \begin{subfigure}[h]{0.48\linewidth}
      \includegraphics[width=1\linewidth]{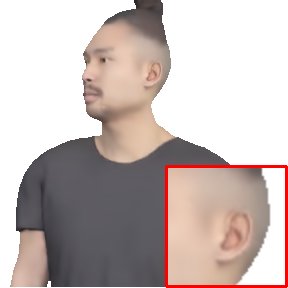}
      \centering
      \caption{Ours 96 kbps}
    \end{subfigure}
    
      \caption{Novel view rendering results with signal compressed at lower bit-rates.}
      \label{fig:render_people}
      \vspace{-5mm}
\end{figure}


\subsection{Qualitative Results}

As shown by the R-D curves, our coding scheme is capable of adapting to different bandwidth constraints. In a low bandwidth scenario, we are interested in the video quality from a nearby novel view point.
With more bandwidth resources, we evaluate our methods in a novel view that is further different from the original ones.

\subsubsection{Limited Bit-Rate Results}

Fig.~\ref{fig:render_people} shows the novel view rendering results from both the synthetic and the real-captured dataset. Comparing (a) and (c), we can see that at a lower bit-rate the proposed scheme does not produce artifacts in the area of the subject's ear. Comparing (b) and (d), our method better reproduce the clothes texture.

\subsubsection{Moderate Bit-Rate Results}
Fig.~\ref{fig:render_people_pcloud} shows qualitative results in higher bit-rates. We render the decoded RGB-D frames from a novel viewpoint further away from the original one. As shown, our method better preserves the geometry and delivers better rendering quality.



\subsubsection{High Resolution}
We mainly conduct the experiments on videos of resolution $288 \times 512$. Since the pre- and post-processor are fully convolutional neural networks, it can be extended to higher resolutions. In this section we present results on the higher-resolution synthetic test dataset, with videos of size $576 \times 1024$. We train another set of pre- and post-processors using $256 \times 256$ crops and calculate $L_{\text{warp}}$ considering the absolute position of the crops. The per-sequence BD-Rate is shown in Table~\ref{tab:highres}.


\begin{table}%
\caption{Per-sequence BD-Rate on the synthetic testing dataset in resolution $576 \times 1024$.}
\label{tab:highres}
\centering
\begin{minipage}{0.45\columnwidth}
\begin{center}
    \footnotesize
\begin{tabular}{c|c}
  \toprule
  Sequence ID & BD-Rate \\
  \midrule
    0 & -31.8 \% \\
    1 & -39.5 \% \\
    2 & -26.5 \% \\
    3 & -37.4 \% \\
    4 & -16.2 \% \\
    \bottomrule
\end{tabular}
\end{center}
\bigskip

\end{minipage}\centering
\begin{minipage}{0.45\columnwidth}

\begin{center}
    \footnotesize
    \centering
\begin{tabular}{c|c}
  \toprule
  Sequence ID & BD-Rate \\
  \midrule
    5 & -27.1 \% \\
    6 & -29.5 \% \\
    7 & -13.0 \% \\
    8 & -24.5 \% \\
    9 & -25.4 \% \\
    \bottomrule
\end{tabular}
\end{center}
\bigskip\centering

\end{minipage}
\vspace{-10mm}
\end{table}%



\section{Conclusion}
In this paper, we propose to wrap a video codec with neural pre- and post-processor, forming a \textit{sandwiched} codec, to upgrade a 2D system to support stereoscopic video conferencing. The neural networks are trained with the help of a JPEG proxy, and are shown to generalize well to \textit{sandwich} sophisticated video codecs like H264 and HEVC. To optimize for the rendering quality under different bit-rate constraints, we propose a simple yet effective warping based distortion measurement for the rate-distortion optimization. We also specifically design the preprocessor for better performance. Experimental results show that our scheme, trained on a synthetic dataset, delivers 29.3\% bit-rate saving over the baseline on both synthetic and real testing videos, and produce better 3D visual representations at a lower bit-rate. Our work has the potential to make stereoscopic teleconferencing more practical and affordable, and we believe it is a promising step towards the future of 3D video communication.

{
    \small
    \bibliographystyle{ieeenat_fullname}
    \bibliography{main}
}

\section{Appendix}

\subsection{Architecture Details}

\begin{table*}[h]
\centering
\caption{Architecture details for the U-Net used in our method.}
\label{tab:unet}
\begin{tabular}{c|c|c|c|c|c|c}
\toprule
Layer  & Type & Input Source & Input Shape & Output Shape & Kernel & Stride \\
\midrule
1 & Convolution & - & (12, H, W) & (64, H, W) & 3x3 & 1 \\
2 & Convolution & 1 & (64, H, W) & (64, H, W) & 3x3 & 1 \\
3 & Convolution & 2 & (64, H, W) & (128, H/2, W/2) & 5x5 & 2 \\
4 & Convolution & 3 & (128, H/2, W/2) & (128, H/2, W/2) & 3x3 & 1 \\
5 & Convolution & 4 & (128, H/2, W/2) & (256, H/4, W/4) & 5x5 & 2 \\
6 & Convolution & 5 & (256, H/4, W/4) & (256, H/4, W/4) & 3x3 & 1 \\
7 & Convolution & 6 & (256, H/4, W/4) & (512, H/8, W/8) & 5x5 & 2 \\
8 & Convolution & 7 & (512, H/8, W/8) & (512, H/8, W/8) & 3x3 & 1 \\
9 & Convolution & 8 & (512, H/8, W/8) & (512, H/16, W/16) & 5x5 & 2 \\
10 & Convolution & 9 & (512, H/16, W/16) & (512, H/16, W/16) & 3x3 & 1 \\
11 & Transposed Conv. & 10 & (512, H/16, W/16) & (512, H/16, W/16) & 5x5 & 2 \\
12 & Concatenation & 11, 8 & 2 x (512, H/8, W/8) & (1024, H/8, W/8) & - & - \\
13 & Convolution & 12 & (1024, H/8, W/8) & (512, H/8, W/8
) & 3x3 & 1 \\
14 & Transposed Conv. & 13 & (512, H/8, W/8) & (256, H/4, W/4) & 5x5 & 2 \\
15 & Concatenation & 14, 6 & 2 x (256, H/4, W/4) & (512, H/4, W/4) & - & - \\
16 & Convolution & 15 & (512, H/4, W/4) & (256, H/4, W/4) & 3x3 & 1 \\
17 & Transposed Conv. & 16 & (256, H/4, W/4) & (128, H/2, W/2) & 5x5 & 2 \\
18 & Concatenation & 17, 4 & 2 x (128, H/2, W/2) & (256, H/2, W/2) & - & - \\
19 & Convolution & 18 & (256, H/2, W/2) & (128, H/2, W/2
) & 3x3 & 1 \\
20 & Transposed Conv. & 19 & (128, H/2, W/2) & (64, H, W) & 5x5 & 2 \\
21 & Concatenation & 20, 2 & 2 x (64, H, W) & (128, H, W) & - & - \\
22 & Convolution & 21 & (128, H, W) & (64, H, W) & 3x3 & 1 \\
23 & Convolution & 22 & (64, H, W) & (12, H, W) & 3x3 & 1 \\
\bottomrule

\end{tabular}
\end{table*}

\begin{table*}
\centering
\caption{Architecture details for the multi-layer perceptron (MLP) used in our method.}
\label{tab:mlp}
\begin{tabular}{c|c|c|c|c|c|c}
\toprule
Layer  & Type & Input Source & Input Shape & Output Shape & Kernel & Stride \\
\midrule
1 & Convolution & - & (C, H, W) & (512, H, W) & 1x1 & 1 \\
2 & Convolution & 1 & (512, H, W) & (C, H, W) & 1x1 & 1 \\
\bottomrule
\end{tabular}
\end{table*}

In this section, we present more details on the neural network architecture used in our method. As mentioned in the main paper, our method is based on the sandwiched codec, which has a neural preprocessor and a postprocessor. Both neural processors are constructed using the U-Net architecture~\cite{ronneberger2015u} and the multi-layer perceptron (MLP) architecture. As shown in Fig.~3 from the main paper, the preprocessor is composed of one U-Net (see architecture details in Table~
\ref{tab:unet}) and two MLPs (see architecture details in Table~\ref{tab:mlp} with $C = 6$ for each of them). The post-processor is composed of one U-Net (see architecture details in Table~\ref{tab:unet}) and one MLP (see architecture details in Table~\ref{tab:mlp} with $C = 12$).

\subsection{Visualization of Neural Codes}

\begin{figure*}[t]
  \begin{subfigure}[h]{0.32\linewidth}
    \centering
    \includegraphics[width=1\linewidth]{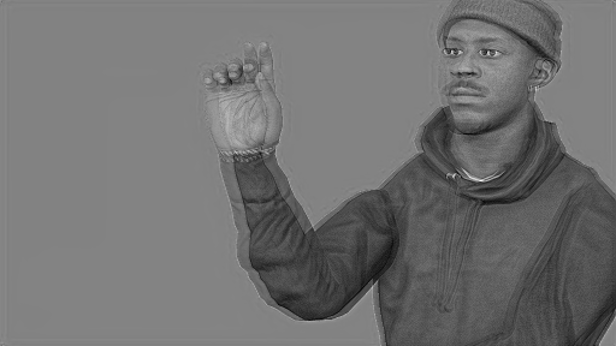}
    \caption{$Y_1'$}
  \end{subfigure}
  \begin{subfigure}[h]{0.32\linewidth}
    \centering
    \includegraphics[width=1\linewidth]{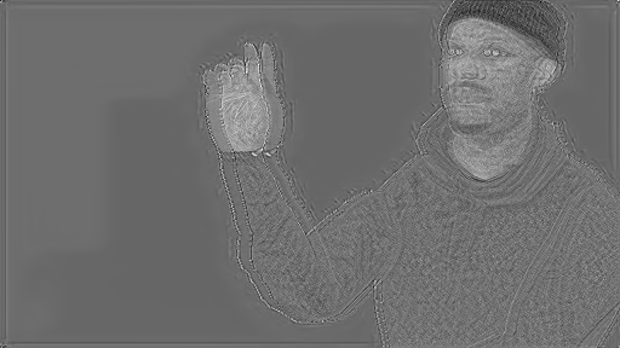}
    \caption{$U_1'$}
  \end{subfigure}
  \begin{subfigure}[h]{0.32\linewidth}
    \centering
    \includegraphics[width=1\linewidth]{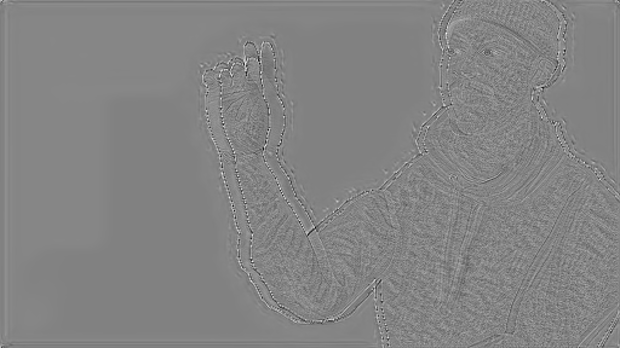}
    \caption{$V_1'$}
  \end{subfigure}

  \begin{subfigure}[h]{0.32\linewidth}
    \centering
    \includegraphics[width=\linewidth]{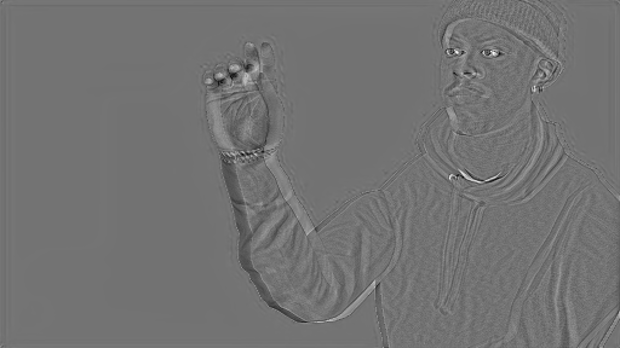}
    \caption{$Y_2'$}
  \end{subfigure}
  \begin{subfigure}[h]{0.32\linewidth}
    \centering
    \includegraphics[width=\linewidth]{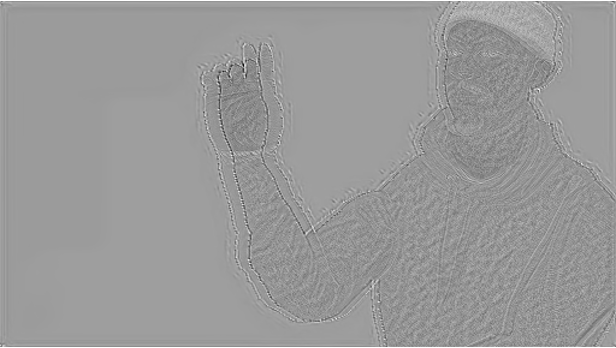}
    \caption{$U_2'$}
  \end{subfigure}
  \begin{subfigure}[h]{0.32\linewidth}
    \centering
    \includegraphics[width=\linewidth]{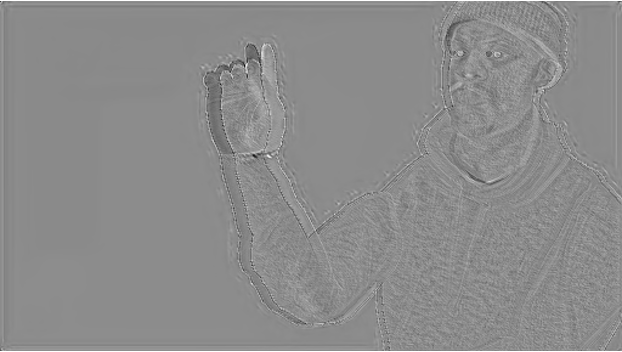}
    \caption{$V_2'$}
  \end{subfigure}

  \begin{subfigure}[h]{0.32\linewidth}
    \centering
    \includegraphics[width=\linewidth]{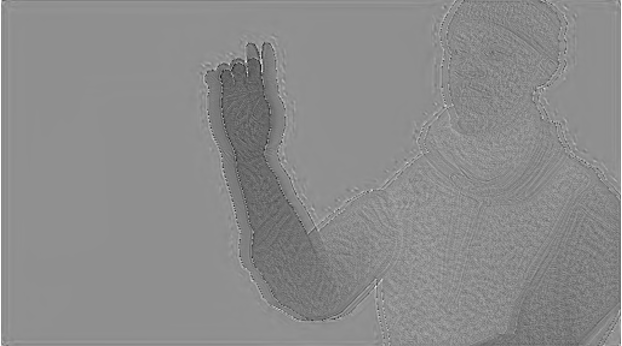}
    \caption{$Y_3'$}
  \end{subfigure}
  \begin{subfigure}[h]{0.32\linewidth}
    \centering
    \includegraphics[width=\linewidth]{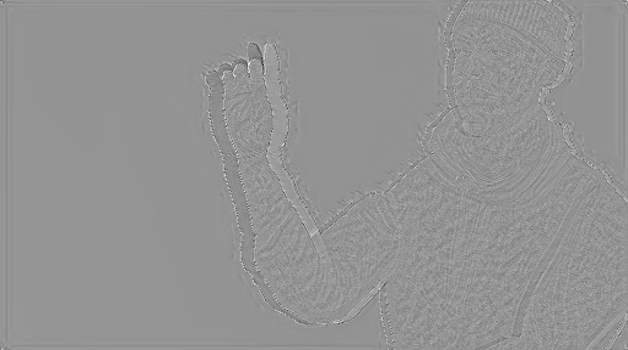}
    \caption{$U_3'$}
  \end{subfigure}
  \begin{subfigure}[h]{0.32\linewidth}
    \centering
    \includegraphics[width=\linewidth]{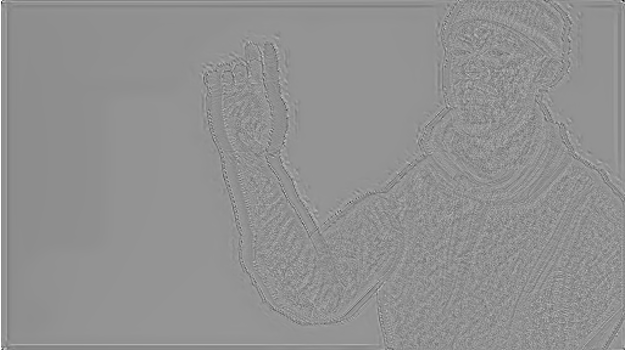}
    \caption{$V_3'$}
  \end{subfigure}

  \begin{subfigure}[h]{0.32\linewidth}
    \centering
    \includegraphics[width=\linewidth]{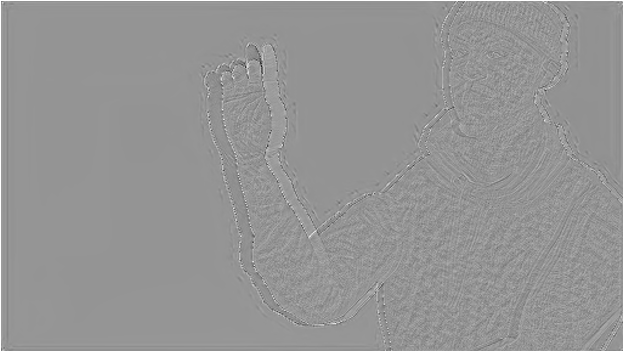}
    \caption{$Y_4'$}
  \end{subfigure}
  \begin{subfigure}[h]{0.32\linewidth}
    \centering
    \includegraphics[width=\linewidth]{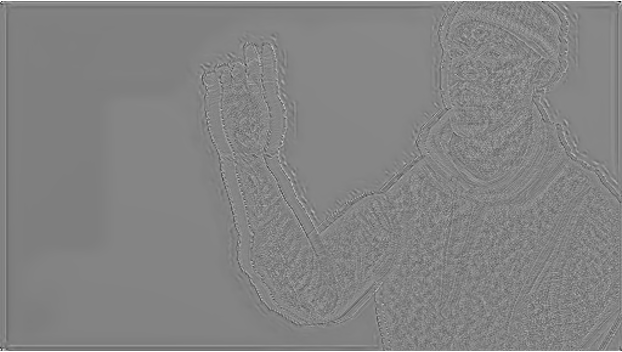}
    \caption{$U_4'$}
  \end{subfigure}
  \begin{subfigure}[h]{0.32\linewidth}
    \centering
    \includegraphics[width=\linewidth]{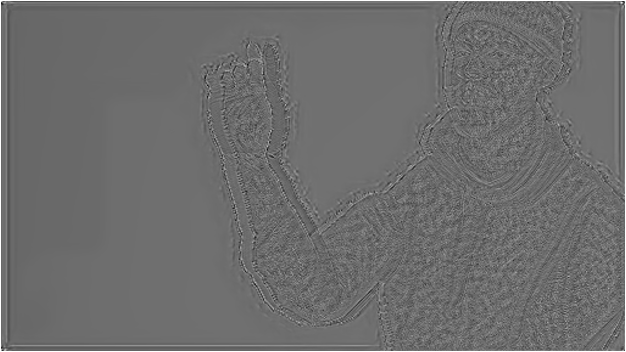}
    \caption{$V_4'$}
  \end{subfigure}

  \caption{Visualization of the latent channels generated by the preprocessor with a low bit-rate setting.}
  \label{fig:latent_low}
\end{figure*}

\begin{figure*}[t]
  \begin{subfigure}[h]{0.32\linewidth}
    \centering
    \includegraphics[width=1\linewidth]{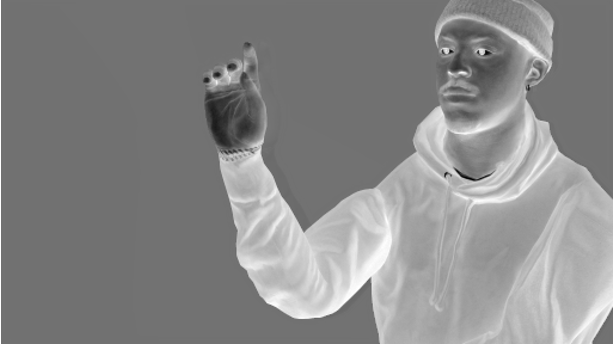}
    \caption{$Y_1'$}
  \end{subfigure}
  \begin{subfigure}[h]{0.32\linewidth}
    \centering
    \includegraphics[width=1\linewidth]{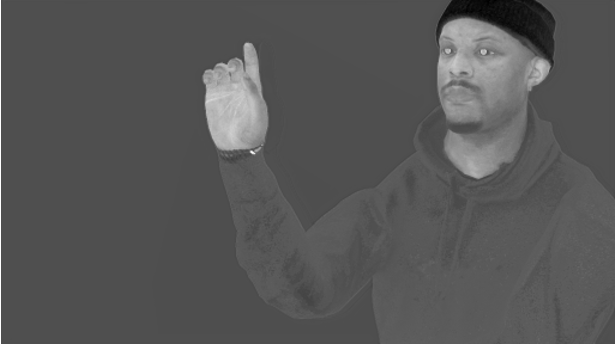}
    \caption{$U_1'$}
  \end{subfigure}
  \begin{subfigure}[h]{0.32\linewidth}
    \centering
    \includegraphics[width=1\linewidth]{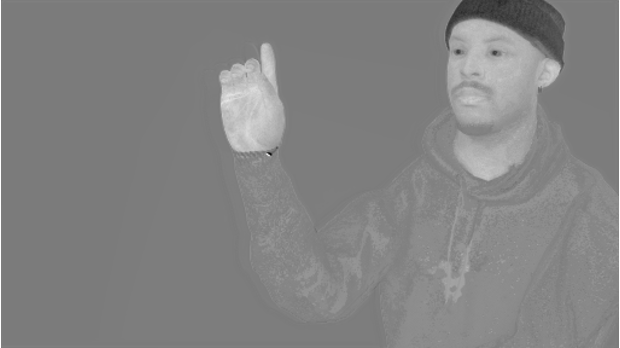}
    \caption{$V_1'$}
  \end{subfigure}

  \begin{subfigure}[h]{0.32\linewidth}
    \centering
    \includegraphics[width=1\linewidth]{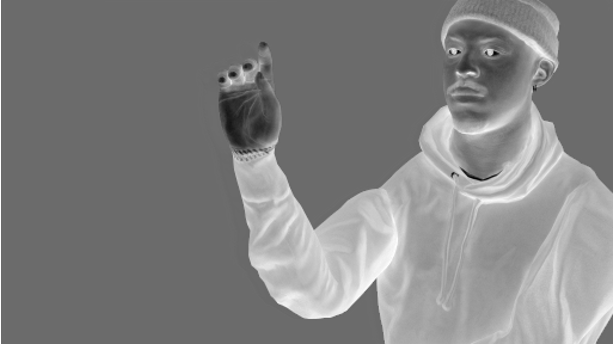}
    \caption{$Y_2'$}
  \end{subfigure}
  \begin{subfigure}[h]{0.32\linewidth}
    \centering
    \includegraphics[width=1\linewidth]{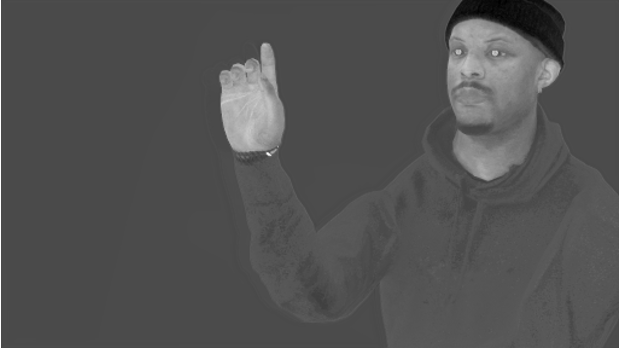}
    \caption{$U_2'$}
  \end{subfigure}
  \begin{subfigure}[h]{0.32\linewidth}
    \centering
    \includegraphics[width=1\linewidth]{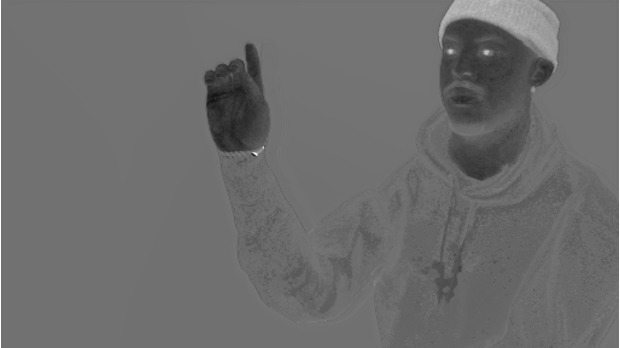}
    \caption{$V_2'$}
  \end{subfigure}

  \begin{subfigure}[h]{0.32\linewidth}
    \centering
    \includegraphics[width=1\linewidth]{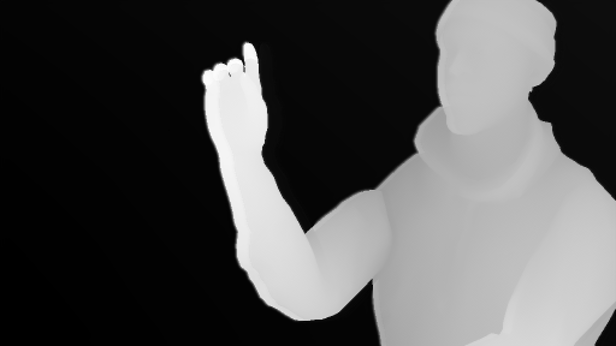}
    \caption{$Y_3'$}
  \end{subfigure}
  \begin{subfigure}[h]{0.32\linewidth}
    \centering
    \includegraphics[width=1\linewidth]{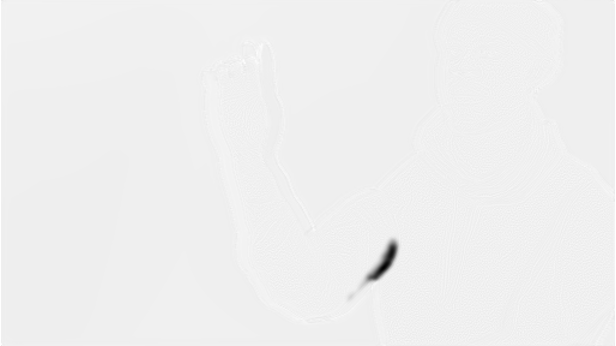}
    \caption{$U_3'$}
  \end{subfigure}
  \begin{subfigure}[h]{0.32\linewidth}
    \centering
    \includegraphics[width=1\linewidth]{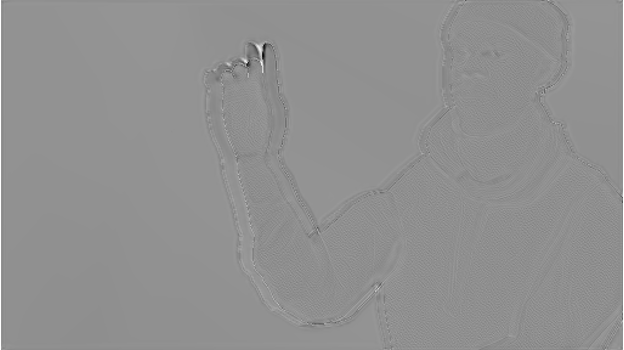}
    \caption{$V_3'$}
  \end{subfigure}

  \begin{subfigure}[h]{0.32\linewidth}
    \centering
    \includegraphics[width=1\linewidth]{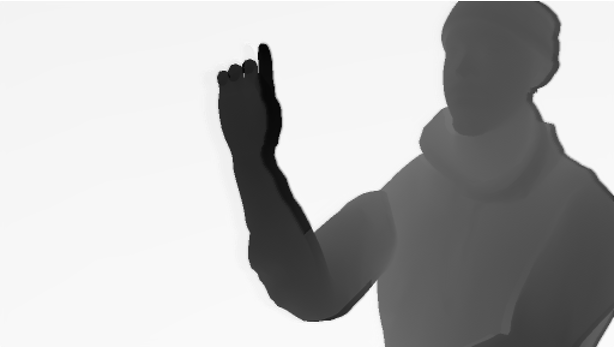}
    \caption{$Y_4'$}
  \end{subfigure}
  \begin{subfigure}[h]{0.32\linewidth}
    \centering
    \includegraphics[width=1\linewidth]{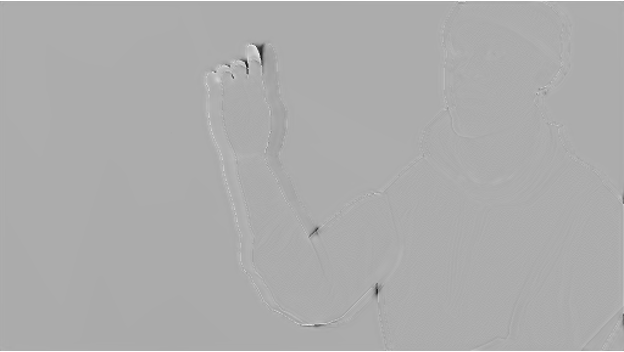}
    \caption{$U_4'$}
  \end{subfigure}
  \begin{subfigure}[h]{0.32\linewidth}
    \centering
    \includegraphics[width=1\linewidth]{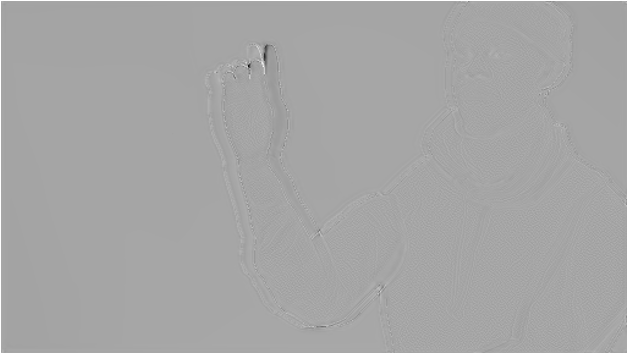}
    \caption{$V_4'$}
  \end{subfigure}

  \caption{Visualization of the latent channels generated by the preprocessor with a high bit-rate setting.}
  \label{fig:latent_high}
\end{figure*}

We visualize the learned neural code generated by the preprocessor, with a low-bit-rate model (in Fig.~\ref{fig:latent_low}) and a high-bit-rate model (in Fig.~\ref{fig:latent_high}), respectively. As described in the main paper, we organize the 12 latent channels into 4 groups, each group containing 3 channels, corresponding to the $Y'$, $U'$, and $V'$ channels of the YUV color space. As shown, at higher bit-rates, the preprocessor learns to maintain more detailed information by fully utilizing the 12 neural code channels. In contrast, at lower bit-rates, the preprocessor learns to compress the information by compressing the input stereo RGB-D information into fewer channels and produce high frequency modulation in the compact latent channels. The resulting neural code channels are thus more sparse and different from the human perceptable RGB-D information.

\subsection{Extension: Relightability}

\begin{figure*}[t]

  \begin{subfigure}[h]{0.06\linewidth}
    \centering
    \footnotesize
    Method
  \end{subfigure}
  \begin{subfigure}[h]{0.2\linewidth}
  \centering
    \footnotesize
  Normal Map
  \end{subfigure}
  \begin{subfigure}[h]{0.2\linewidth}
  \centering
    \footnotesize
  Relighted
  \end{subfigure}
  \begin{subfigure}[h]{0.06\linewidth}
    \centering
    \footnotesize
    Method
  \end{subfigure}
  \begin{subfigure}[h]{0.2\linewidth}
  \centering
    \footnotesize
  Normal Map
  \end{subfigure}
  \begin{subfigure}[h]{0.2\linewidth}
  \centering
    \footnotesize
  Relighted
  \end{subfigure}
  
  \begin{subfigure}[h]{0.06\linewidth}
    \centering
    \footnotesize
    Ground\\Truth
  \end{subfigure}
  \begin{subfigure}[h]{0.2\linewidth}
    \centering
    \includegraphics[width=1\linewidth]{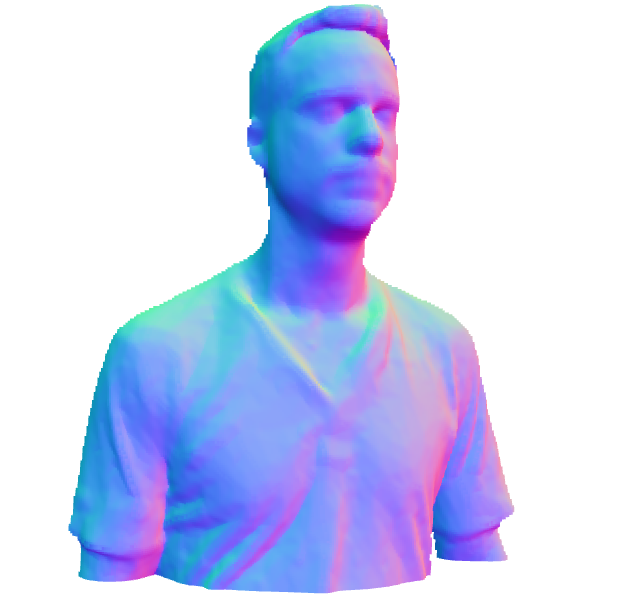}
  \end{subfigure}
  \begin{subfigure}[h]{0.2\linewidth}
    \centering
    \includegraphics[width=1\linewidth]{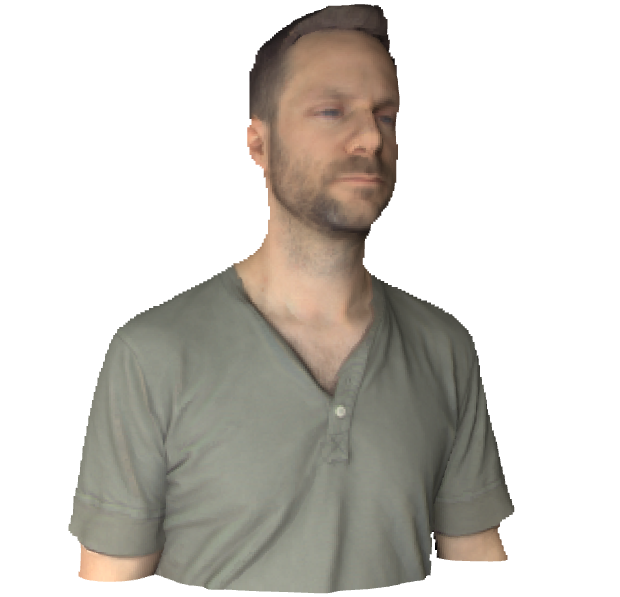}
  \end{subfigure}
  \begin{subfigure}[h]{0.06\linewidth}
    \centering
    \footnotesize
    Ground\\Truth
  \end{subfigure}
  \begin{subfigure}[h]{0.2\linewidth}
    \centering
    \includegraphics[width=1\linewidth]{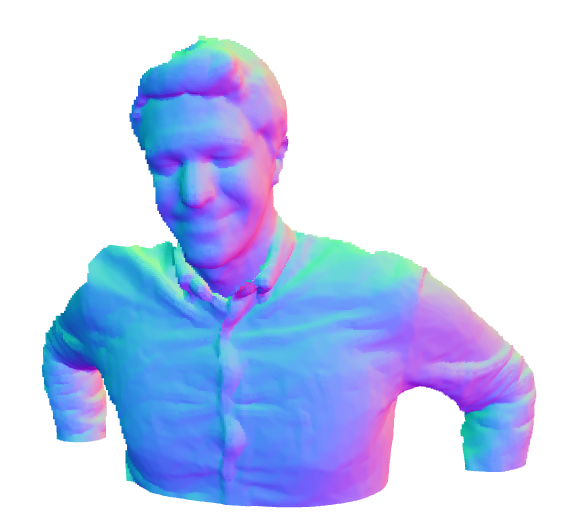}
  \end{subfigure}
  \begin{subfigure}[h]{0.2\linewidth}
    \centering
    \includegraphics[width=1\linewidth]{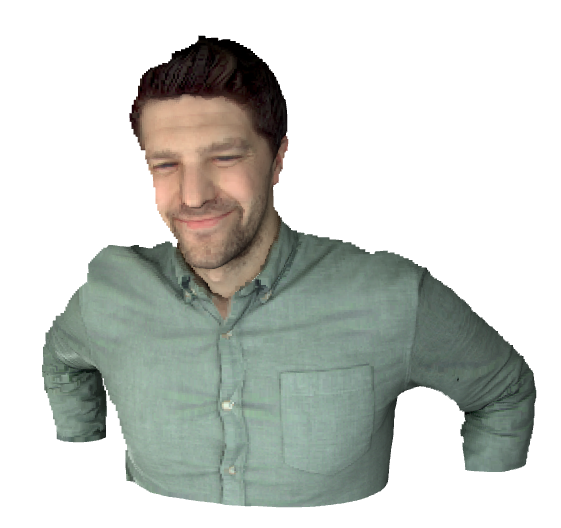}
  \end{subfigure}
  
  \begin{subfigure}[h]{0.06\linewidth}
    \centering
    \footnotesize
    H264 \\
    0.78 mbps
  \end{subfigure}
  \begin{subfigure}[h]{0.2\linewidth}
    \centering
    \includegraphics[width=1\linewidth]{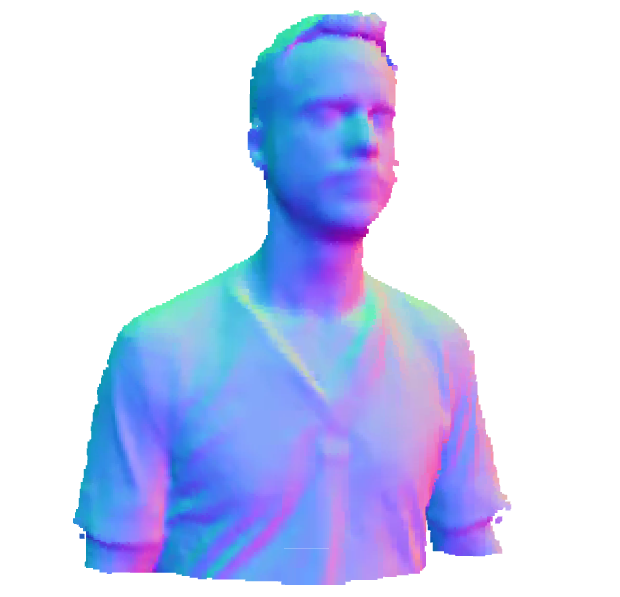}
  \end{subfigure}
  \begin{subfigure}[h]{0.2\linewidth}
    \centering
    \includegraphics[width=1\linewidth]{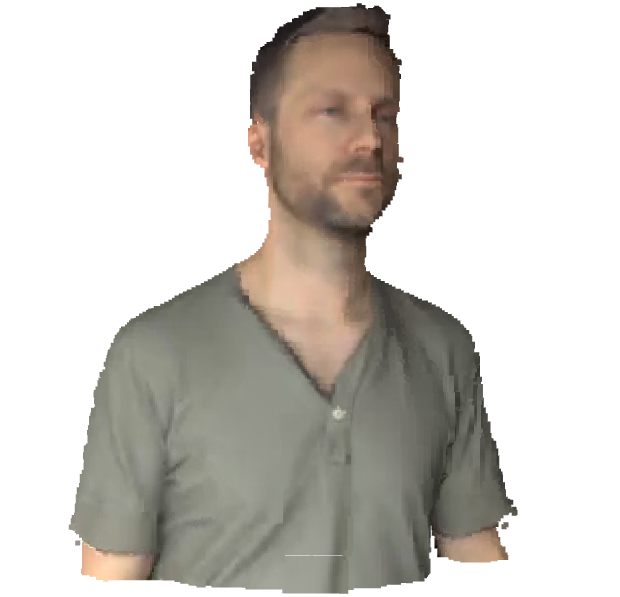}
  \end{subfigure}
  \begin{subfigure}[h]{0.06\linewidth}
    \centering
    \footnotesize
    H264 \\
    1.46 mbps
  \end{subfigure}
  \begin{subfigure}[h]{0.2\linewidth}
    \centering
    \includegraphics[width=1\linewidth]{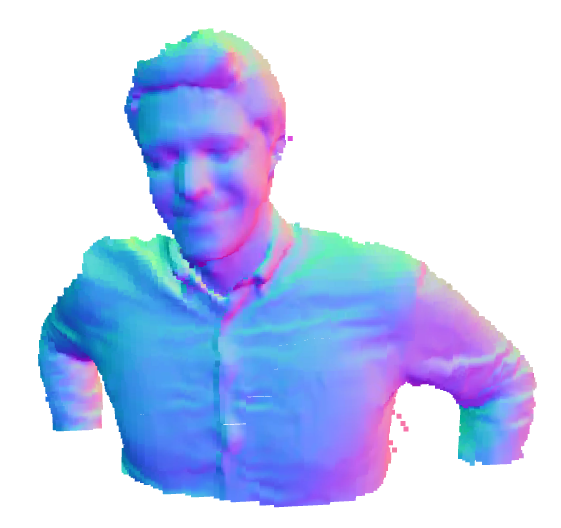}
  \end{subfigure}
  \begin{subfigure}[h]{0.2\linewidth}
    \centering
    \includegraphics[width=1\linewidth]{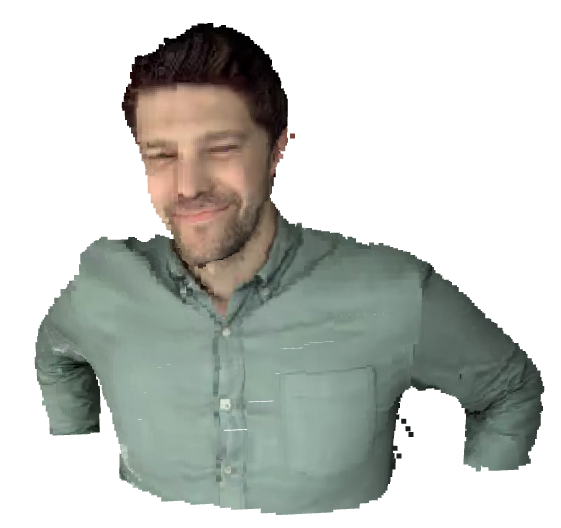}
  \end{subfigure}
  
  \begin{subfigure}[h]{0.06\linewidth}
    \centering
    \footnotesize
    Ours \\
    0.76 mbps
  \end{subfigure}
  \begin{subfigure}[h]{0.2\linewidth}
    \centering
    \includegraphics[width=1\linewidth]{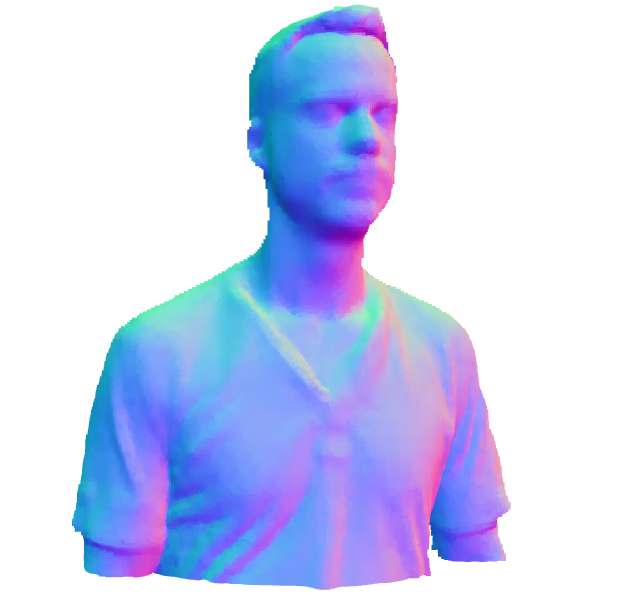}
  \end{subfigure}
  \begin{subfigure}[h]{0.2\linewidth}
    \centering
    \includegraphics[width=1\linewidth]{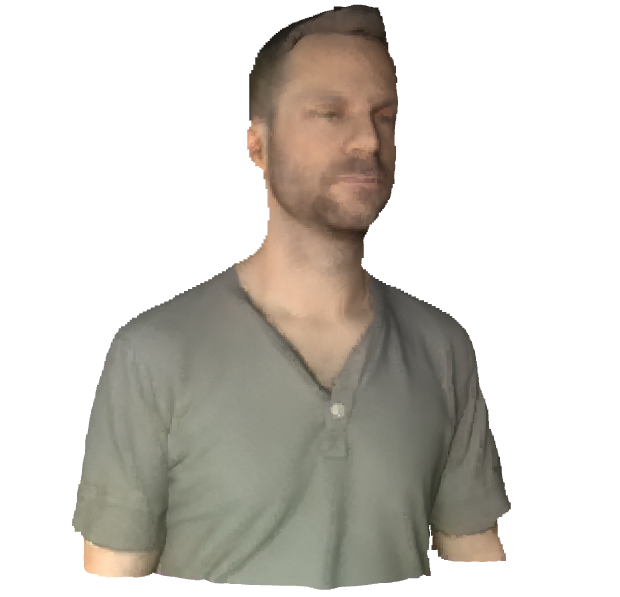}
  \end{subfigure}
  \begin{subfigure}[h]{0.06\linewidth}
    \centering
    \footnotesize
    Ours \\
    1.40 mbps
  \end{subfigure}
  \begin{subfigure}[h]{0.2\linewidth}
    \centering
    \includegraphics[width=1\linewidth]{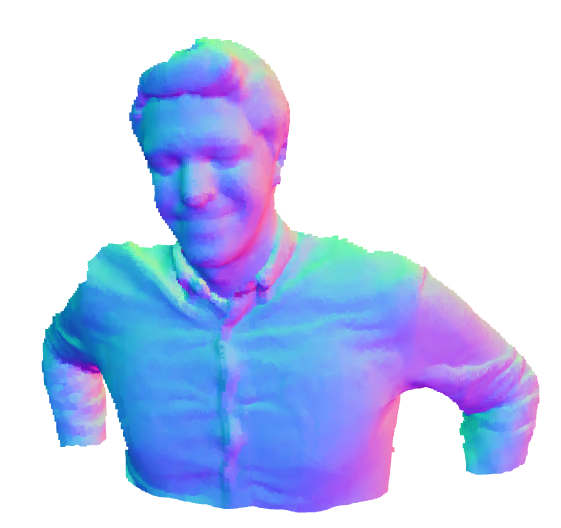}
  \end{subfigure}
  \begin{subfigure}[h]{0.2\linewidth}
    \centering
    \includegraphics[width=1\linewidth]{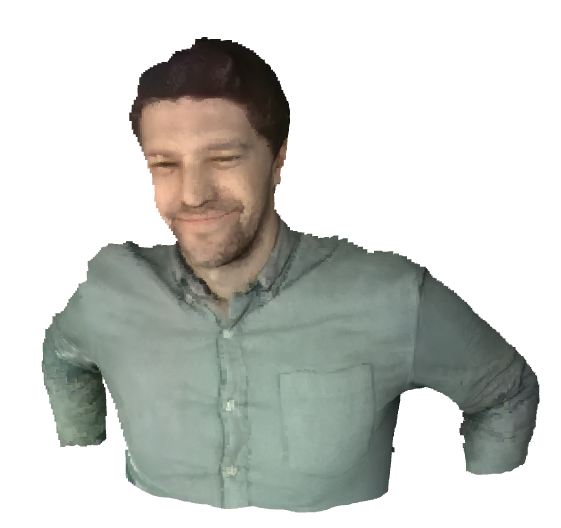}
  \end{subfigure}
  
    \caption{Qualitative results with \textit{The Relightable} dataset showing the relightability of the decoded signal in the extension setting.}
    \label{fig:relight}
  \end{figure*}

Since decoder side relightability is sometimes desired in a scene transmission pipeline. In this section, we demonstrate a simple extension of our method to support decoder-side relighting by transmitting normal maps together with the RGB-D signal in our sandwiched codec. We qualitatively evaluate the extension setting by relighting the transmitted 3D representation. Results shown in Fig.~\ref{fig:relight} demonstrate that our method can be easily extended to a relighting rendering pipeline and maintains better performance.


\end{document}